# To be or not to be: a translation reception study of a literary text translated into Dutch and Catalan using machine translation[1]

**Ana Guerberof-Arenas and Antonio Toral**

**University of Groningen**

*Abstract*

This article presents the results of a study involving the reception of a fictional story by Kurt Vonnegut translated from English into Catalan and Dutch in three conditions: machine-translated (MT), post-edited (PE) and translated from scratch (HT). 223 participants were recruited who rated the reading conditions using three scales: Narrative Engagement, Enjoyment and Translation Reception. The results show that HT presented a higher engagement, enjoyment and translation reception in Catalan if compared to PE and MT. However, the Dutch readers show higher scores in PE than in both HT and MT, and the highest engagement and enjoyments scores are reported when reading the original English version. We hypothesize that when reading a fictional story in translation, not only the condition and the quality of the translations is key to understand its reception, but also the participants' reading patterns, reading language, and, perhaps language status in their own societies.

---

*Introduction*

As part of a larger research project[2] that involves the exploration of creativity and the use of machine translation (MT) in the production of a translated literary text, this article details the results of exploring translation reception when mediated by technology, in our case neural MT.

Translation reception is an under-explored area in literary translation and further research is highly needed (Walker 2021) to understand translation from a wider perspective and its impact on the user, but also to understand how new tools in the translation process might affect the final users of translation. Although there is some published work that considers the translation of literary texts versus the original texts (H. Kruger 2013; Walker 2020), the impact MT has on the literary translation process and, subsequently, on readers has not been thoroughly explored yet.

In our previous work (Guerberof-Arenas and Toral 2020), we selected a short story, *Murder in the Mall* (Nuland 1995), to compare different translation modalities using a Narrative engagement scale (Busselle and Bilandzic 2009), enjoyment questions (Hakemulder 2004) and a translation reception questionnaire. The methodology used was partially borrowed from Mangen and Kuiken (2014) that compared engagement between on-line and on-paper reading. Our results showed that translation from scratch (HT) scored higher for creativity than post-editing (PE) and MT, and that HT scored higher in narrative engagement and translation reception and slightly lower than PE in enjoyment while MT showed statistically significant lower results than HT and PE. We hypothesized that this difference between HT and PE was due to a higher creativity score in HT and, therefore, a different source text with more creative challenges for translators would increase the gap between the conditions if reception was to be considered.

---

[2] CREAMT project https://cordis.europa.eu/project/id/890697

With this hypothesis in mind, we have expanded the cohort of participants to 223, included another language combination, English to Dutch, and, more importantly, chosen a text in which, unlike our pilot experiment, style prevails over action.

*Related Work*

As mentioned in the introduction, there is existing work that considers the translation of literary texts versus the original using an eye-tracker (H. Kruger 2013; Walker 2020). Kruger analysed domesticating and foreignizing translation strategies by readers of translated children's books. She looked at dwell time, fixation count, first-fixation duration, and glances count for areas of interest that reflect these strategies. Walker looked at the salient features of literature (foregrounding) to see if readers' cognitive effort is equivalent between source text and target text by using gaze duration and total fixation duration. He found that these salient features result in higher cognitive effort and greater diversity in readers' visual attention. Although these studies do not deal directly with MT or PE, they signal the relevance of looking at salient features in combination with users' responses.

The Digital Opinions on Translated Literature project, DIOPTRA-L, (Kotze et al. 2021) looked at the salience of the fact of translation in readers'opinions: what are the main concepts, and emotional and evaluative parameters, for those readers who leave their opinions in Goodreads. They found that the fact of translation is not particularly salient for these readers and, in fact, when translation is mentioned, it tends to accompany a poor review. They speculate that translation might act here as a "scapegoat" for readers' preferences. Although it varies by language combination, translation is more salient for those that read a translation into English than into other languages.

In the pilot experiment mentioned in the introduction (Guerberof-Arenas and Toral 2020), the narrative engagement, enjoyment and translation reception was measured in a cohort of 88 Catalan readers and differences were found depending on the condition. HT scores higher in narrative engagement and translation reception and is slightly lower than PE in enjoyment. However, there are no statistically significant differences between HT and PE for any of these variables. MT, unsurprisingly, has the lowest engagement, enjoyment and translation reception scores. However, certain categories in the scale such as attentional focus, emotional engagement and narrative presence do not show statistically significant differences across the conditions. This could be related to the nature of the story readers were presented with, where action was more relevant than style. It is also clear from the results that readers enjoyed PE marginally more or as much as HT, even if they appreciate that HT was a better translation.

Colman et al. (2021) reported preliminary results of an eye-tracking study in which participants read a full novel in Dutch, alternating between MT and the published translation. They compared the reading process of participants reading both versions and found increased fixation, gaze duration, and regression in the MT version, but not in the saccade amplitude (a quick movement of the eyes between fixations), although there is an increase in word-skipping. This study focuses on overall reading effort and error typologies, however; it does not analyse any specific creative traits or the impact of post-edited texts on the reader.

In a recent conference paper, Whyatt, Lehka-Paul and Tomczak (2022) carried out a translation reception study with 20 students that read a marketing text using an eye-tracker to correlate translation effort, the presence of errors and the easiness of text production with the gaze data and hence reading effort by participants. They found that high quality translations are easier to read but that quality does not seem to affect the comprehension of a text nor the participants' decision to buy a product. This exploratory study is a part of larger research called Read Me project which looks at the relationship between the translation process and the reading experience.

Stasiomiti and Sosoni (2022) examine the reception of marketing and literary texts in the English to Greek language combination with 100 readers. They found that the post-edited creative texts do not seem to differ from the HT texts in the number of errors and in the feedback from the readers. However, these translations were created by translation students some of which were active translators, but no professional literary translators were engaged in the experiment.

There has been more active research in reception of audiovisual translation gaining prominence with the work of D'Ydewalle and colleagues since the 1980s (1984; 1989; 1987). J.L. Kruger and Orrego-Carmona offer a detailed overview that covers how viewers behave when reading subtitles (e.g., analysing standard, intralingual and reversed subtitling), presentation of subtitles, experience with humour, accessibility studies, audio-description, and dubbing versus subtitling (2018; 2018). Relevant to this article, Ortíz Boix (2016) examined two conditions (HT and PE) in the translation of wildlife documentaries.

The results of a panel of experts and 56 end-users established no significant differences between the two conditions. Hu et al. (2021) looked at the reception of subtitles in massive open online courses from English to Chinese. They found that viewers showed higher reception scores in the PE condition, though the difference with MT was not statistically significant. Viewers appreciated the HT condition, but it did not generate the highest scores. The eye-tracking data does not clearly show if one condition requires more cognitive effort than the other.

*Overall Reading Experience: Methodology*

The CREAMT project is articulated in two main axes. Results on the first one (about creativity) were reported in a recent article (Guerberof-Arenas and Toral 2020). We are presenting here the results of our second axis driven by the research question:

> **RQ$_1$**: Do users reading translated material produced using different translation conditions have different reading experiences?
>
>> **RQ$_{1.1}$**: And does the readers' experience vary between languages? In our case, Dutch and Catalan.

To answer these questions, the following experimental setting was used.

*Source Text*

The chosen literary text had to meet the following conditions: a) higher creative potential than *Murder in the Mall*, b) not present in the data used to train the NMT engines, c) engaging enough to measure reading experience in a wider audience, d) short enough to be read in approximately thirty minutes, e) not too outdated so all generations could engage with it and f) a text that would not infringe copyright laws.

We selected a Kurt Vonnegut story from the Project Gutenberg Literary Archive

Foundation.[3] 2BR02B was published in January 1962 in the digest magazine *Worlds of If Science Fiction* and it was later included in the collection *Bagombo Snuff Box* (Vonnegut 1999). This story is set in a futuristic world where death has been eradicated and people live forever, the only caveat is that for one person born, another one must die. Vonnegut centres the story on several characters, a father to be, a doctor, a government official, a hospital orderly and a painter. These characters exemplify different attitudes towards the world they are immersed in. Vonnegut creates a new world and, thus, invents terms, roles, institutions and expressions to describe it.

The text contains 123 paragraphs, 234 segments and 2,548 words. The text has a Flesch Reading Ease index of 79 and a Flesch-Kincaid Grade Level of 4.6. The readability indexes indicate that the text is not an overly difficult text to read for an English speaker. To our knowledge, this story has not been translated and published into Catalan nor Dutch.

*Target Texts*

The short story was translated from English into Dutch (NL) and Catalan (CA) into three conditions: a translation done from sratch (HT), post-edited MT output (PE) and the raw MT output. These languages were selected for two reasons: convenience of sample and availability of two MT engines already trained and tested at University of Groningen. The HT and PE versions were provided by four professional translators who specialize in literary translation.[4]

To reduce the effect of the translator in the experiment, i.e. the risk that a reader would engage more with one translator's work because they preferred that translator's style, each professional translated and post-edited fifty percent of the text, and then the text was aggregated so, in fact, each of the two translation conditions, HT and PE, contained the aggregated translations of these two translators per language.

---

[3] https://www.gutenberg.org/files/21279/21279-h/21279-h.htm

[4] To recruit the translators, two databases were consulted (Expertisecentrum Literair Vertalen and Associació d'escriptors en llengua catalana). Some translators recommended others that were contacted based on availability. The translators used in the first experiment were recruited as reviewers for Catalan.

These translations were reviewed by five professional translators. The reviewers were unanimous in raking HT as an Extremely good translation, MT as an Extremely bad translation, and PE as a Neither good nor bad translation (Guerberof-Arenas and Toral 2022). MT contains more errors than the other two conditions put together, PE also contains a strikingly higher number of errors than HT (double the errors in CA and three times in NL), even if the same two translators worked on the HT and PE. Further, the HT has the highest number of creative shifts followed by PE and lastly by MT.

The target texts presented to the participants in the current study are the translated texts not including the changes from reviewers.

*Reading Conditions*

Unknown to them, all readers were randomly assigned a condition. READINGA corresponded to the PE version, READINGB to the MT version and READINGC to the HT version. Since in the Netherlands is not common to read English-speaking authors in translation, we decided to add an additional condition READINGD, English condition, to the NL group. Therefore, in the end, each CA reader was assigned one out of three conditions and each NL reader one out of four. In this article, the name of the condition is used for ease of understanding.

*Questionnaire*

Two on-line questionnaires were distributed to participants using Qualtrics,[5] one translated to Catalan and the other to Dutch. The participants first read the information brochure and consent form and if they decided to participate they were taken to these sections sequentially:

1. *Demographics and Reading Patterns.* This section contains 13 items on demographics and reading patterns (e.g. "What genre do you usually read? How often have you read in Catalan/Dutch in the last 12 months?").

2. *Comprehension Questions*. After reading the text, the participants answered 10 four-choice questions to ensure basic comprehension of the story. The participants could only continue if five questions were answered correctly. Even though, it would have been interesting to analyse the full comprehension (1 to 10) depending on the condition, this section intended to act as a filter for spurious responses as the questionnaire was posted in a variety of on-line reading sites.

3. *Narrative Engagement.* Participants were presented with a 12-item Narrative engagement scale (Busselle and Bilandzic 2009) with a 7-point Likert-type responses. The questionnaire focuses on four categories: Narrative understanding (e.g. "At points, I had a hard time making sense of what was going on in the story"), Attentional focus (e.g. "While reading, I found myself thinking about other things"), Narrative presence (e.g. "The story created a new world, and then that world suddenly disappeared when the story ended"), and Emotional engagement (e.g. "I felt sorry for some of the characters in the story").

---

[5] The questionnaire and the anonymised data are available at https://github.com/AnaGuerberof/CREAMT.

4. *Visual Imagery*. We added three questions for mental imagery (e.g. "When I was reading the story I had an image of the main characters in mind") borrowed from the Story world absorption scale (Kuijpers et al. 2014) to complement with more literary items the previous scale designed for media studies.

5. *Enjoyment*. Participants were then asked to answer a 3-item scale to address enjoyment: "How much did you enjoy reading the text?", "Do you think this text is an example of good literature?", "Would you recommend this text to a friend?". These questions were borrowed from experimental research on the effects of foregrounding[6] in relation to specific text qualities (Dixon et al. 1993; Hakemulder 2004).

6. *Translation Reception.* To our knowledge, there is no existing scale to measure translation reception. Therefore, we devised a scale with 11 items and 7-point Likert-type responses (e.g. "How easy was the text to understand?", "I thought the text was very well written", "I found words, sentences or paragraphs that were difficult to understand").

7. *Debriefing and Payment Questions*. At the end of the questionnaire, the participants were debriefed about the nature of the research. Only then were they informed about the author and the nature of the text. Following this, they were asked how much money they would be willing to pay for this translation or original piece and, if their condition was MT, to rate the quality of the engine.

---

[6] In literary studies, foregrounding refers to the effect of certain language features that serve to change the attention of the reader.

*Experimental Workflow*

Figure 1 summarises in a sequential flow the steps the participants followed:

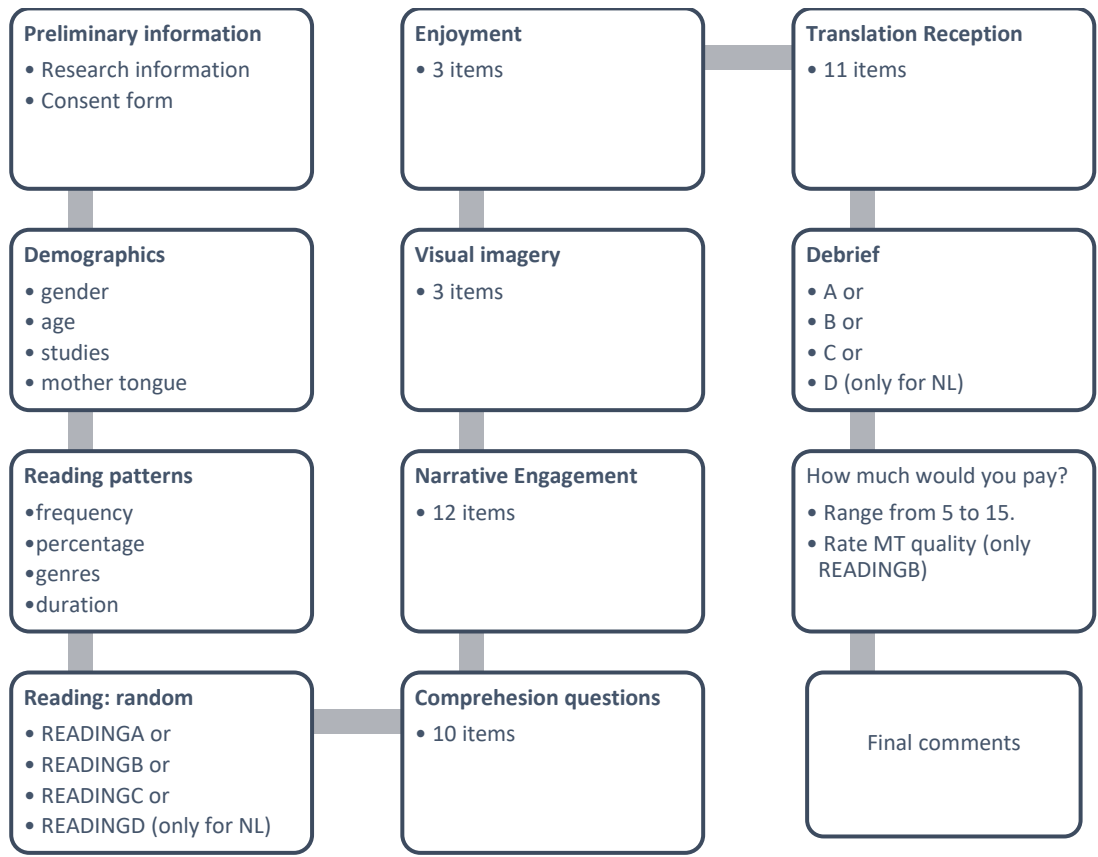

Figure 1. *Experimental Workflow*

*Preliminary Analysis: Participants*[7]

The criteria for the inclusion of participants was that they were native CA and NL speakers and 18 years or older. Readers were offered 10-euro compensation for their time. To recruit participants that fit the profile (frequent readers of fiction) forums for readers were targeted to avoid any bias towards translation scholars, practitioners or students, but without necessarily excluding this group of readers.

---

[7] All our statistical analysis is available at https://github.com/AnaGuerberof/CREAMT

The experiment was advertised in the following sites: Goodreads (Lectura en català Group,[8] Fanatieke Nederlandse Lezers,[9] Netherlands Flanders Group),[10] Hebban,[11] Relats en català (a web forum for readers and writers in Catalan),[12] Catalans in Ireland,[13] Catalans in Holland,[14] Samenlezenisleuker,[15] Leesclub In de boekenkast (Facebook groups),[16] reading clubs organized by libraries in Catalonia[17] and in the Netherlands such as Senia[18] or Literaire Studentenvereniging Flanor,[19] and writing schools such as Escola d'Escriptura de l'Ateneu Barcelonès, and Dutch speaking members of the EACWP.[20] To obtain more participants, group of students studying Catalan and Dutch philology were targeted through different professors at Universitat Oberta de Catalunya, Universitat Rovira i Virgili and University of Groningen. In turn, some participants received the link to our study from other participants through private messaging. The advertisements were posted at different intervals during the time the questionnaire was active, 7th September 2021 to 21st January 2022. Figure 2 shows the distribution of participants according to their provenance (N = 223).

---

[8] https://www.goodreads.com/group/show/61003-lectura-en-catal
[9] https://www.goodreads.com/group/show/79675-fanatieke-nederlandse-lezers
[10] https://www.goodreads.com/group/show/223-netherlands-flanders-group
[11] https://www.hebban.nl/community
[12] http://relatsencatala.cat/
[13] https://www.facebook.com/groups/CatalansIrlanda
[14] https://www.facebook.com/groups/203190096419372
[15] https://www.facebook.com/groups/451488498379185
[16] https://www.facebook.com/groups/1574870836130714
[17] Biblioteques.cat
[18] https://www.senia.nl/pages/Senia/Home
[19] https://www.flanor.nl/en/interested
[20] European Association of Creative Writing Programmes

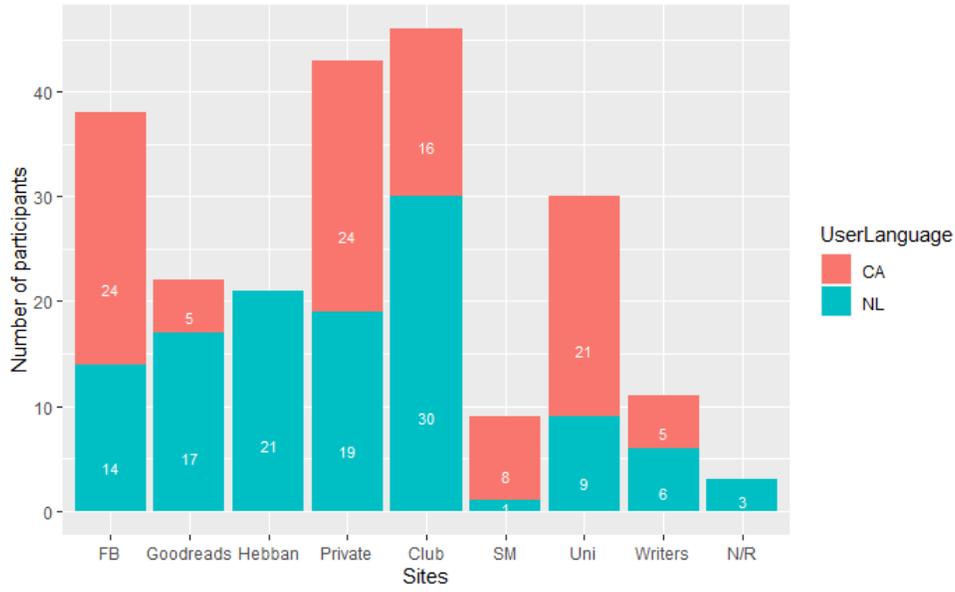

Figure 2. Distribution of participants according to sites and mother tongue

The questionnaire was completed by 103 CA and 120 NL participants. Table 1 contains the participants' demographic information.

Table 1. Participant's Demographics

| Categories | | | | | |
|---|---|---|---|---|---|
| Gender | Women | Men | Non-binary | Prefer not to say | Total |
| | 168 | 48 | 4 | 3 | 223 |
| Age | 18-34 | 35-54 | 55-74 | 75 or older | Total |
| | 94 | 63 | 58 | 8 | 223 |
| Studies | Secondary | Some college | College degree | Professional | Total |
| | 30 | 27 | 151 | 15 | 223 |
| Mother tongue | Catalan | Catalan/Spanish | Dutch | Dutch/English | Total |
| | 50 | 53 | 118 | 2 | 223 |

| Profession | Language related | Other | | Total |
|---|---|---|---|---|
| | 53 | 165 | | 218 (5 unanswered) |

*Similarities and Differences between Groups*

Although advertisements were published in similar sites, some characteristics varied within the group of final participants. In both languages most participants are women (68 out of 104 in CA and 100 out of 120 in NL) with a college degree (151 out of 223). However, the NL group has a higher percentage of participants with a college degree than the CA group (72.5% and 62%, respectively). Regarding age, the CA readers have a higher percentage of participants in the age bracket 35-54 while the NL readers have a higher percentage in the age bracket 55-74.

The NL and CA participants' reading patterns are also quite different. The following questions about their reading patterns were posed to the participants and they had to rank them on a 5-point Likert scale: "How often have you read a book in the last two years?",[21] "How much do you like reading?",[22] "How often have you read a book in CA/NL in the last two years?", [23] "How long did you spend reading on these occasions?".[24]

The variable *Reading_Patterns* represents the average given to these four questions. Figure 3 shows the results for the two languages (N = 223).

---

[21] 1= Never, 2= Once every three months, 3= Once a month, 4= Once or twice per week, 5= Daily
[22] 1 = Dislike a great deal, 2 = Dislike somewhat, 3 = Neither like or dislike, 4 = Like somewhat, 5 = Like a lot
[23] 1= Never, 2= Once every three months, 3= Once a month, 4= Once or twice per week, 5= Daily
[24] 1= I don't read, 2 = less than 15 minutes, 3 = between 15 and 30 minutes, 4 = between 31 and 60 minutes, 5 = more than 60 minutes

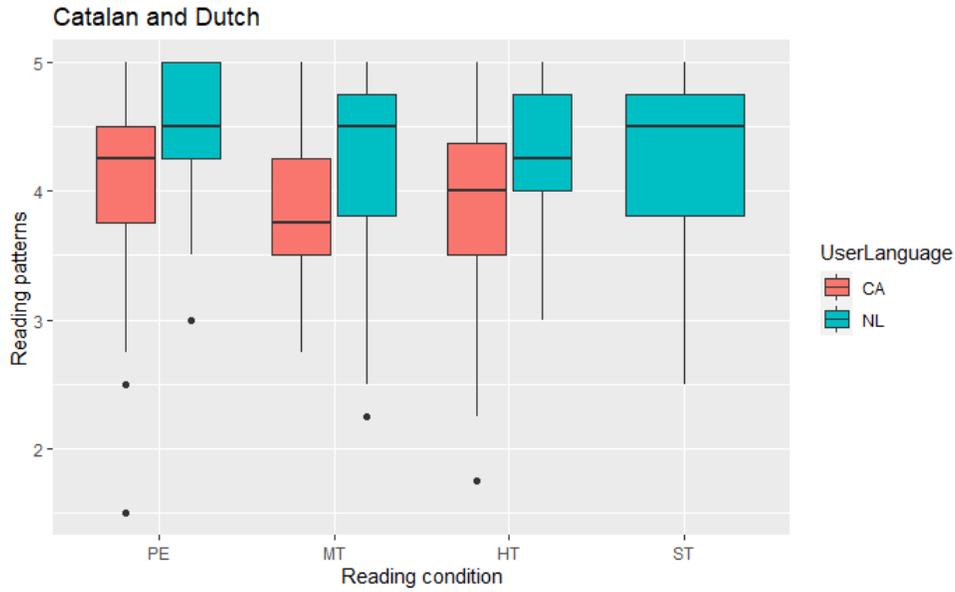

Figure 3. Reading patterns by language and condition

NL participants have higher *Reading_Patterns* than the CA readers in all conditions. Since the data is not normally distributed, a Wilcoxon rank sum test was run to see if there were significant differences between the two groups. The measured median reading pattern for CA (4) was statistically significantly lower than that for NL (4.5) with a moderate effect ($p < .001$, effect size $r = 0.33$). The reasons for this could be the number of people that responded to our survey according to their age and their provenance (different sites). For example, Figure 2 shows that the NL group has more respondents from reading clubs and from reading sites (such as Hebban and Senia) while the CA has a higher number from universities and Facebook groups. There were no statistically significant differences between the variable *Reading_Patterns* and the three translation conditions, so participants with a higher or lower reading pattern were distributed evenly among our conditions.

In conclusion, most of our readers are women with a high-level education. The NL readers are slightly more educated, older and declare having higher reading patterns than the CA group.

*Comprehension Questions*

All participants in the study had to answer at least 5 questions correctly out of 10 to be able to continue. The mean value for all participants is above 7 for all reading conditions. The highest value is for condition ST (M = 8.3), followed by HT (M = 8.2), PE (M =7.9) and MT (M = 7.8). There are no significant differences between the conditions overall. However, a Kruskal-Wallis H test for non-parametric data[25] on the CA responses indicates that there is a significant difference between conditions (H(2) = 6.33, $p < .05$). Post-hoc comparisons using the Conover test with the Holm-Bonferroni correction show statistically significant differences between PE and HT (Z = -2.46; $p < .05$). There are no significant differences among the NL readers. Therefore, the comprehension questions are impacted by the reading condition only for CA: the HT readers show a significantly higher comprehension. To have a deeper understanding of comprehension, it could be interesting to let participants continue regardless of the number of correct questions, but this would certainly jeopardize the validity of the experiment since the data was collected on-line.

---

[25] This test is used to determine if there are statistically significant differences between two or more groups within the independent variable (reading condition) when the scale uses rank-based nonparametric values.

*Narrative Engagement*

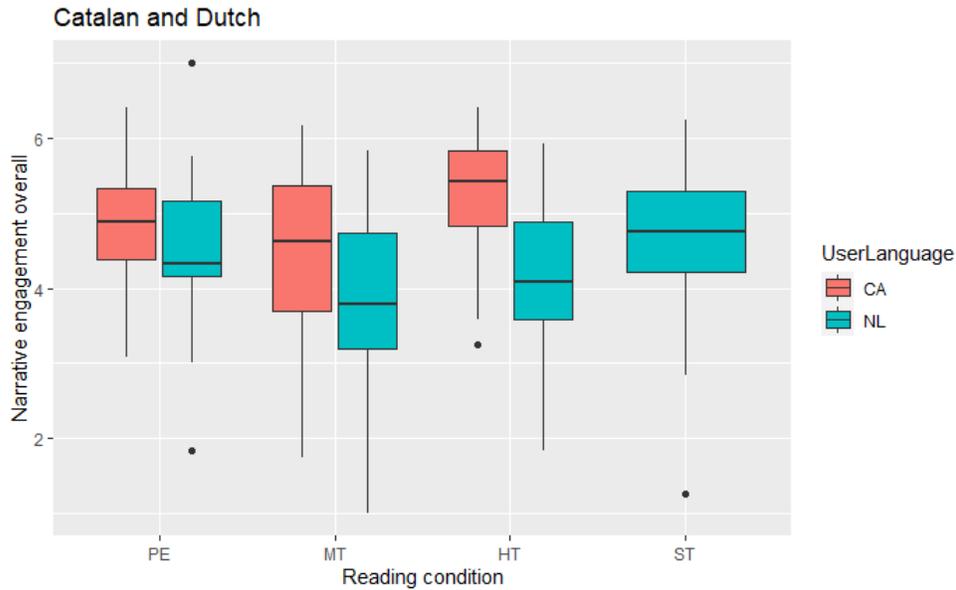

Figure 4 shows the results of the average ratings given for the 12-item scale in a 7-point Likert scale for the two languages (N = 223). The Cronbach's alpha reliability[26] coefficient (α) is 0.85 for all the items in the scale, which is considered a reliable score.

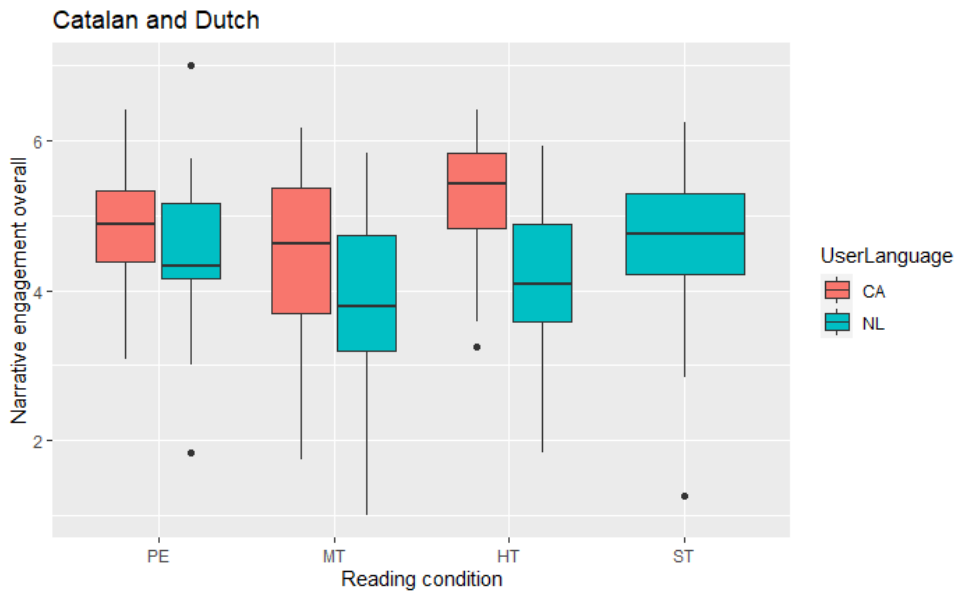

Figure 4. Narrative engagement overall per language and condition

---

[26] Cronbach's alpha measures the internal consistency of a scale. It gives an idea of how the items are interrelated and measure similar concepts.

The CA participants show a higher engagement than NL participants in all conditions. CA HT has the highest score, but this is not replicated in NL where ST, followed by PE, shows the highest scores. This is odd since reviewers indicated that HT was a translation of higher creativity than PE in both languages, and reviewers were unanimous on their assessment of quality of these two conditions (Guerberof-Arenas and Toral 2022). It appears that NL readers do not share the same opinion as professional reviewers.

Since the data does not meet the assumptions for a linear regression model that would allow us to check the interaction between engagement, condition and languages, and given that the groups have different reading patterns and the participants read different texts, we decided to analyse the two language groups separately.

The variable *Narrative_Engagement* was explored according to the reading condition using the Kruskal-Wallis H test for non-parametric data. In CA there is a statistically significant difference between conditions (H(2) = 11.90, p < .001) with a mean rank score of 65.64 for HT, 48.28 for PE and 41.68 for MT. Post-hoc comparisons using Conover-Iman test with the Holm-Bonferroni correction show statistically significant differences between PE and HT (Z = -2.54; p = .01) and between MT and HT (Z= -3.51; p = .00). For NL, there is a statistically significant difference between conditions (H(3) = 10.12, p = .02) with a mean rank score of 72.88 for ST, 65.57 for PE, 58.27 for HT and 45.52 for MT. Post-hoc comparisons show statistically significant differences only between MT and ST (Z= -3.15; p = .02)[27]. This is indeed surprising. It appears that in NL there is no significant difference between the translation conditions.

To clarify these findings and following the same statistical methods, each category in the

---

[27] The p values presented here are not adjusted. In the Conover-Iman test, the null hypothesis is rejected if p <= alpha/2.

Narrative engagement scale was analysed per language. Table 2 shows the results together with their significance.

Table 2. Narrative engagement results per category and language

| Category | CA | NL |
| --- | --- | --- |
| Narrative understanding<br>3 items | Significant<br>MT ≠ HT Z = -3.26, p = .01 | Significant<br>MT ≠ ST Z= -4.11, p = 0 |
| Attentional focus<br>3 items | Significant<br>MT ≠ HT, Z= -2.65, p =.02 | Not significant |
| Narrative presence<br>3 items | Not significant | Not significant |
| Emotional engagement<br>3 items | Not significant | Not significant |
| Overall Narrative engagement<br>12 items | Significant<br>PE ≠HT (Z = -2.54; p = .01)<br>MT ≠ HT (Z= -3.51; p = .001) | Significant<br>MT ≠ ST (Z= -3.15; p = .01) |

The main problem with the texts translated using MT when it comes to engagement seems to be in the category *Narrative understanding*, which relates to the ease of comprehension of a story. The participants answered to these questions: "At points, I had a hard time making sense of what was going on in the story", "My understanding of the characters is unclear", "I had a hard time recognizing the thread of the story". Therefore, and even if these readers did respond satisfactorily to the comprehension questions, they did not perceive this activity as an easy one in the MT condition. In CA, there is also an issue with *Attentional focus*, the state of being engaged and not distracted. The participants reacted to these statements: "I found my mind wandering while reading the story", "While reading, I found myself thinking about other things", "I had a hard time keeping my mind on the story". We had suspected that the MT readers would find it more difficult to be transported to the story, because they would find elements of distraction that would prevent this. This seems to be the case only for CA readers.

This is partially in line with our previous finding (Guerberof-Arenas and Toral 2020) where *Narrative understanding* was the category most affected by the reading condition. It seems that when looking at a translated text, *Narrative presence* (the feeling that one has entered the world of the story) and *Emotional engagement* (feeling for and with the characters) are categories stronger linked to the story, and, hence, the world created by the writer rather than to the translation.

*Visual Imagery*

This section of the questionnaire was added to see if depending on the condition, readers have a higher level of difficulty when imagining the characters, the situations, and the world depicted in the story. We found minor differences (none significant) in these scores between conditions, although the NL readers, as in the rest of the survey, score these statements lower than the CA readers.

*Enjoyment*

Figure 5 shows the results of the average scores given for this 3-item scale ($\alpha = 0.87$) in a 7-point Likert scale for the two languages (N = 223).

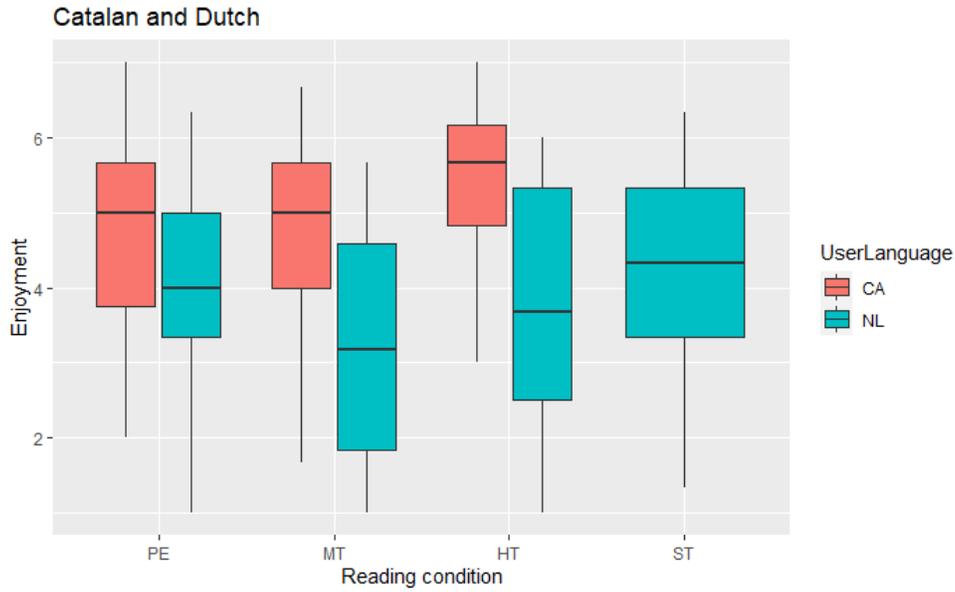

Figure 5. Enjoyment per language and condition

NL readers enjoy the story less than CA readers. We see similar patterns as in the Narrative engagement section, namely CA readers had the highest enjoyment in HT and NL readers in ST. As before, a Kruskal-Wallis test was carried out to compare enjoyment between the conditions per language group. In CA, there was evidence of a difference (H(2) = 6.65; p < .05) with a mean rank score of 62.63 for HT, 46.68 for PE and 46.49 for MT. Pairwise comparisons, however, show no statistically significant differences between the conditions. In NL, there was no evidence of a significant difference. The median rank scores are 71 for ST, 64.52 for PE, 58.61 for HT and 47.97 for MT. These results differ from our previous experiment (Guerberof-Arenas and Toral 2020) where MT scored significantly lower in enjoyment than PE and HT.

*Translation Reception*

Figure 6 shows the results of the average ratings given for the 8 quantifiable items (α = 0.79) in a 7-point Likert for the two languages (N = 193 since the ST condition is not included).

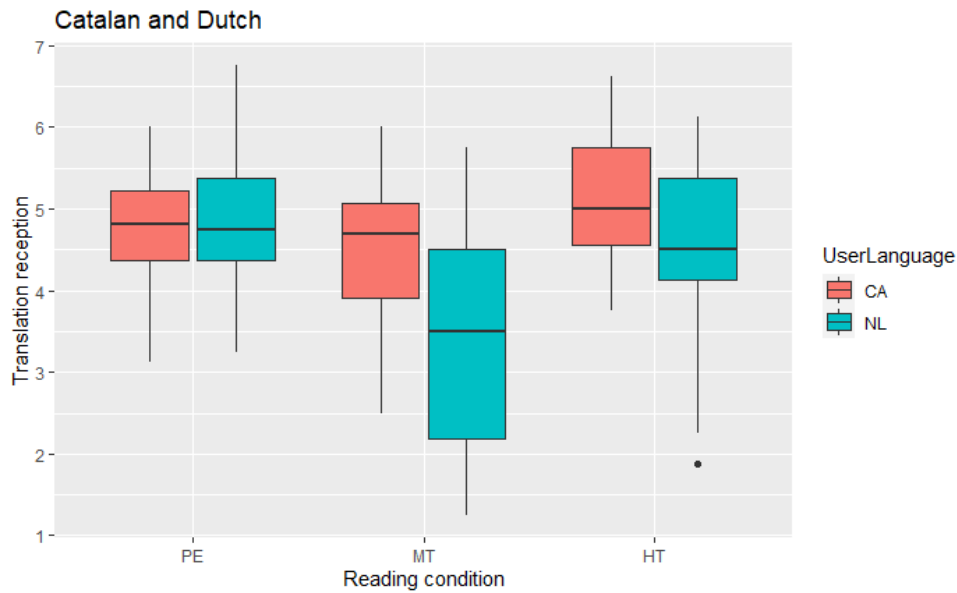

Figure 6. Translation reception per language and condition

NL participants gave lower rates to MT and HT on average, but similar scores to CA participants when rating PE. As with previous variables, in CA the reception is highest for HT, but, unexpectedly, the difference in reception between PE and MT does not seem as pronounced. For NL participants, the MT scores are indeed lower than PE and HT. A Kruskal Wallis H test shows statistically significant difference between conditions in CA ($H(2) = 8.66$, $p = .01$) with a mean score rank of 62.6 for HT, 51.62 for PE and 41.47 for MT. Post-hoc comparisons show statistically significant only between MT and HT ($Z= -3.05$; $p = .01$). For NL, there is a statistically significant difference between conditions ($H(2) = 16.68$, $p = 0$) with a mean rank score of 55.47 for PE, 51.39 for HT and 29.78 for MT. Post-hoc comparisons show statistically significant differences between MT and HT ($Z= -3.55$; $p < .001$), and between PE and MT ($Z= 4.15$; $p < .001$). These results, that appear to favour the CA MT system, could be an indication of the quality of this system.

However, NL readers tended to give lower scores overall, and they reported higher reading patterns, so the results could be an indication of a more demanding reader.

*Comparison between Translation and Original Text*

Participants that read the original English text were also presented with the following questions/statements: "How easy was the text to understand?", "I thought the text was very well written", "I found words, sentences or paragraphs that were difficult to understand", "I found words, sentences or paragraphs that I especially liked" and "Would you like to read a text by the same author". Since the first four were also asked in the translation conditions, they were compared to see if, as with other variables, readers of the original text gave a higher score than those that read in translation. Figure 7 shows the average scores for each condition including the ST.

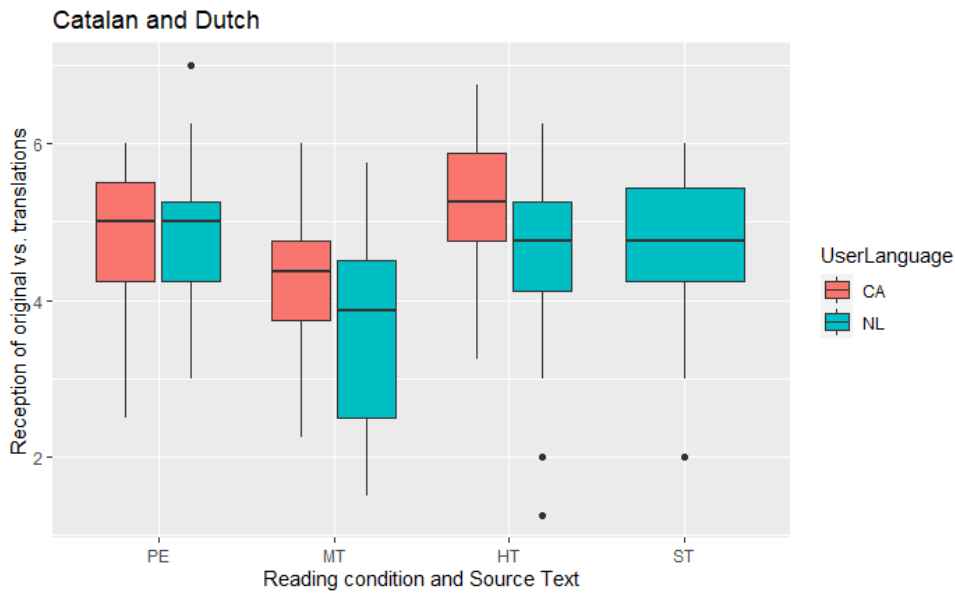

Figure 7. Translation vs Source Text Reception

The ST did not obtain higher scores than the translation conditions, the CA HT condition has the highest scores. Indeed, A Kruskal Wallis test shows statistically significant difference between conditions in CA (H(2) = 15.92, p = 0) with a mean score rank of 65.4 for HT, 53.93 for PE and 36.68 for MT. Post-hoc comparisons show statistically significant only between MT and HT (Z= -4.27; p < .001) and between PE and MT (Z= 2.58; p = .01). For NL, there is a statistically significant difference between conditions (H(3) = 15.48, p = 0) with a mean rank score of 71.82 for PE, 66.83 for ST, 64.15 for HT, and 39.45 for MT. Post-hoc comparisons show statistically significant differences between MT and HT (Z= -2.94; p < .01), between PE and MT (Z= 3.79; p < .001) and between MT and ST (Z= -3.24; p < .001). Again, here we can see that the quality of the CA and NL MT systems plays a role in reception, while there are no significant differences between PE and HT.

*How much would you pay?*

After debriefing the participants about the author and reading condition, they were asked for the amount of money they would be willing to pay: 1 = less than a euro, 2 = between 1 and 5 euros, 3 = between 5 and 10 euros, 4 = between 10 and 15 euros and 5 = more than 15. Figure 8 shows the values obtained.

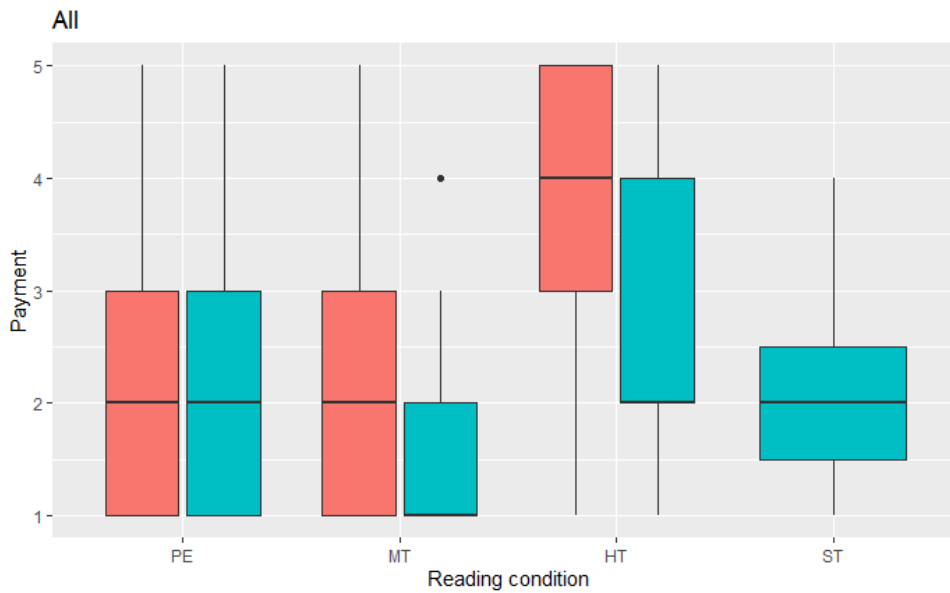

Figure 8. Payment per language and condition

Participants are willing to pay between 1 euro and more than 15 euros when they realize that the translation is done by professional literary translators, and over 15 euros in the case of CA, while they are willing to pay less for PE, between less than 1 euro and 10 euros and even less for MT, between less than 1 euro and 10 euros in CA, and 5 euros in NL. The amount that NL participants are willing to pay for the ST is remarkably low, perhaps since this author is no longer alive, they might be aware that the text is copyright free. We wonder if participants would be willing to pay the same amount if we had asked them before debriefing. At any rate, the results show that participants do place a value in hand-crafted translations, even if they are not willing to pay the same price range for the original piece. A Kruskal Wallis test shows statistically significant difference between conditions in CA (H(2) = 30.24, p = 0) with a mean score rank of 73.25 for HT, 44.40 for PE and 37.74 for MT. Post-hoc comparisons show statistically significant only between PE and HT (Z= -5.22; p = 0), and between MT and HT (Z= -6.10; p = .0). For NL, there is a statistically significant difference between conditions (H(3) = 17.31, p = 0) with a mean rank score of 71.17 for ST, 70.89 for HT, 57.52 for PE and 41.98 for MT. Post-hoc comparisons show statistically significant differences between MT and HT (Z= -4.63; p < .001), and not between the other pairs. Indeed, participants are willing to pay more for the service of professional translators.

*Quality of the MT Systems according to the Readers*

The participants that were randomly assigned to the MT condition were asked after the debrief to indicate their opinion on the quality of the system: 1 (Extremely bad) and 7 (Extremely good). Figure 9 shows the results per language for the variable MTQA (N =

64).

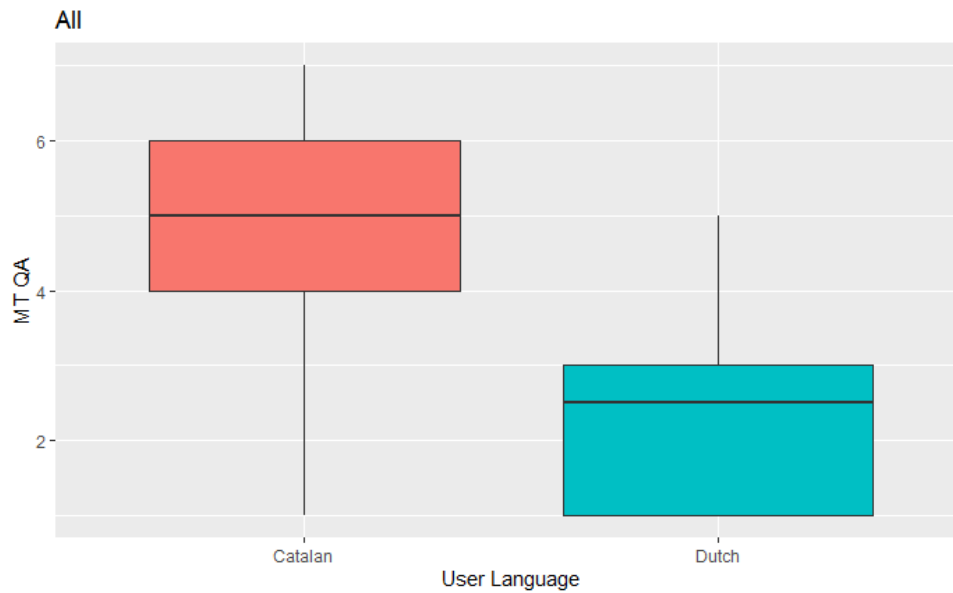

Figure 9. Reported MT Quality per language

CA participants ranked the MTQA noticeably higher than NL participants. Since the data is not normally distributed, a Wilcoxon was used and this showed that the difference is significant with a large effect (p < .001, effect size r = 0.62). The reasons for this could be that NL participants were more demanding of quality since they show higher reading patterns, but it could also indicate differences between the quality of the systems. The NL translators had also reported a lower quality perception than the CA translators (Guerberof-Arenas and Toral 2022).

*Summary of Findings*

Table 3 gives an overview of the quantitative results from the survey distributed per language.

Table 3. Summary of findings

| Category | Catalan | Dutch |
| --- | --- | --- |
| Comprehension | **Highest values for HT** Significance between PE and HT PE ≠ HT (Z = -2.46; p < .05). | **Highest values for ST** No significance between conditions |
| Reading Patterns | No significance between conditions | No significance between conditions Significantly higher values than the CA group p = .00, effect size r = 0.33 |
| Narrative Engagement | **Highest values in HT** PE ≠ HT (Z = -2.54; p = .01) MT ≠ HT (Z= -3.51; p = .00) Significant Higher scores for HT | **Highest values in ST** MT ≠ ST (Z= -3.15; p = .01) Higher scores for ST, not significant |
| Imagery | No significance between conditions | Lower scores than CA, but no significance between conditions |
| Enjoyment | **Highest values for HT** Significance between conditions PE ≠ HT (Z = -2.27; p = .03) MT ≠ HT (Z = -2.29, p = .04) | **Highest values for ST** No significance between conditions. |
| Translation Reception | **Highest values of HT** Significance between conditions MT ≠ HT (Z= -3.05; p = .01) | **Highest values in PE** Significance between conditions MT ≠ HT (Z= -3.55; p = .00) PE ≠ MT (Z= 4.15; p = .00) |
| Translation vs. Original | **Highest values HT** Significance between conditions | **Highest values PE** Significance between conditions |

|  | MT ≠HT (Z= -4.27; p = .00) | MT ≠ HT (Z= -2.94; p = .01) |
|  | PE ≠ MT (Z= 2.58; p = .01) | PE ≠ MT (Z= 3.79; p = .00) |
|  |  | MT ≠ ST (Z= -3.24; p = .00) |
| Payment | **Highest values HT** | **Highest values HT** |
|  | Significance between conditions | Significance between conditions |
|  | PE ≠HT (Z= -5.22; p = .00) | MT ≠ HT (Z= -4.63; p = .00) |
|  | MT ≠ HT (Z= 6.10; p = .00) |  |
| MT Quality | **Mean value 5 (Slightly good)** | **Mean value 2.5 (Bad)** |

*Comments from the Readers*

The readers were asked to make comments about those parts of the text they found difficult, those parts they liked, how they realized they were reading a translation and final comments to complement the quantitative analysis.

*Difficult Parts*

There were no comments from the readers of the original English, while all the other conditions had comments, with MT having the most.

In the HT version, both CA and NL found problems related to the story ("it was difficult to know what was going on at the beginning", "the character descriptions", "the phone numbers", "it was a terribly boring piece"), and certain words in Catalan such as *trigèmins* (triplets), "the use of a very high register mixed with a low register", lack of commas or too much use of the conjunction *ni* (either/neither).

In the PE version, the CA readers found some words confusing (such as the name of one of the characters, Affleck"), [28] the use of unknown vocabulary to them, the sobriquets[29] (e.g., the Dutch *kattenbak* for the English Catbox), the word *zelador* in Catalan (caretaker). Some comments referred to the story: the first part was confusing, one reader complained about the text being too long, a NL reader commented on bizarre and outdated parts of the text.

In the MT version, there were comments about incorrect sentence structures ("Some sentences were built up a bit oddly and referred to each other in unclear terms"), odd punctuation, lack of coherence, incorrect use of verb tenses; the use of the wrong vocabulary such as *foto* (the Dutch word for photography instead of painting), "*Ruimte voor twee* [space for two] was translated as *Kamer voor twee* [a room for two]", words left in English in the CA version such as "daub, sobriquet, sheepdip [sic], drupelets [sic]" . But there were also similar comments about the story itself as in the previous conditions: a confusing beginning, or the use of neologisms. Only one participant referred to machine translation: "It seems like a sort of machine translation; there were words that did not correspond to the meanings that we could get from the context. For example, there was the word *jackrabbit*".

As we have seen before in comments from readers for this type of experiments, when commenting about difficulties, HT and PE readers tend to comment more about intrinsic characteristics of the story, with a few comments on the language, while the MT readers tend to refer mainly to the characteristics of the text, the translation, such as the structure,

---

[28] This name was translated into Catalan to reflect the same image evoked by the name Wehling in English, someone that is suffering or in pain. When compiling the PE version that was post-edited by two translators some decisions were made to unify the version, and the name Affleck was left in this version. The CA readers were not aware of the original name, so it is surprising that they remarked on this name.

[29] Vonnegut creates a series of new terms to refer to the death system based on cultural references. The translators adapted these to the target language.

punctuation, out of context vocabulary and to a sense of general "oddity" that confuses them.

*Preferred Sections*

As to the parts that the readers liked, there were no comments again for the original English, but for all translation conditions in both languages. The number of comments for MT is the highest, which is surprising.

In the HT version, participants refer to parts of the story they had liked: "the painter describing the mural", "The father's moral conflict when deciding on the life or death of his children and the irony of the painter", "The end has undoubtedly been the best part of the story.", "I found the paragraph in which the father took his decision to kill Hitz, Leora and himself quite impressive". There were some references to the translation itself *El número de telèfon CONOC (ser o no ser)* [The telephone number 2BR02B (to be or not to be)], "The synonyms/names of Suicide Studio", "I thought the names for the gassing in connection with birth control were funny, like *kattenbak*".

In the PE version, readers also refer to certain moments in the text such as "The atmosphere was set by the phrases that for every child born, someone else must die", "The description of the painting", "The predominant colour purple", "The bit about the future father taking fate into his hands and making sure his three children are all allowed to live", "The paragraph where he describes the Garden of Eden". Many also refer to the ending of the story, its moral dilemma and the painter's reflections. As before, they also refer to the translations: "The nicknames for the Federal Bureau of Termination, such as *kattenbak*", "the sentence with the flowery descriptions of the body that regulates killing *dompelaar*, *kattenbak*", "I've been amused by some of the names used for the gas chambers. Also, the phone number, which when you read it is 'To be or not to be'".

In the MT version, the readers mentioned certain sections again such as the description of the Garden of Eden, the father's moral dilemma, reflections about getting older, life cycles and general reflections of the characters, and their characterization (sincerity, frankness), and the impressive ending. They also mentioned certain translations, for example, a NL reader mentioned a particular line of the song "*Ik ga van deze oude planeet af, laat een baby ijn [sic] plaats innemen*" [I am leaving this old planet, let a baby take my place] or "I really enjoyed reading words I haven't heard in a long time. One of the reasons I like to read science fiction in Catalan. Also to support the language and writers", "I liked the way the writer/translator was able to divide the scenes using the language", "*La teva ciutat t'ho agraeix però més t'ho agraeixen les futures generacions*" [Your city is grateful to you, but future generations are more grateful to you]. Also, in this version, the readers refer to word choices for the gas chambers and the name of the characters, that in this case were left in the original English.

When describing what they have liked, the readers focused more on what they liked about the story, but also on parts of the translation. This was surprising for MT; it seems that after all MT did capture a certain essence of the text. It seems that when it comes to expressing positive aspects of a story, the reading condition was not as determinant for the readers even if the MT readers did give lower scores to this condition throughout the experiment.

*How did you realize it was a translation?*

Readers reported a higher value (meaning that the fact of the translation was more salient) for MT, then HT and PE. As a follow-up question, we asked them how they arrived at this conclusion (we wanted to know if this was due to errors or to the setting of the story).

In the HT version, most readers commented, as we suspected, that characters' names and the action that occurred in the United States or Chicago, or even the translated names of the offices echoed names in the United States. Some mentioned that some words or structures were not very frequent in their language, or literal, and that the style resembled English with "little flourish". In NL, the title 2BR02B (to be or not to be) was left in English, so this was also mentioned. One NL reader mentioned "I generally prefer not to read translations, but here it wasn't distracting" which reflects a tendency in the Netherlands to read literature from English-speaking authors in the original language, but also an appreciation for this translation by this reader.

In the PE version, readers commented that some expressions, constructions, and words read like translations, or that they were "clumsy", or that the translation was artificial and "cold" but, as before also the characters' names, the story not being set in their country, but in Chicago, or the names of the offices in the NL translation. One NL reader had read the story in English before, so he knew it was a translation when he started reading.

In the MT version, the readers made more comments about the style, incoherent sentences, wrong sentence structure, words out of context, absence of certain pronouns, unnatural dialogues, expressions translated literally from English without corresponding to the concept, for example, in Catalan *ningú a les meves sabates* [nobody in my shoes], punctuation, grammatical errors, alternate use of the treatment pronouns, terms left in English, confusing story thread. One reader also mentioned "Phrases a bit unconnected, but since it was so surreal, I didn't know if it was because of that or because of a bad translation". As before, they also commented on the style of narration with few adjectives and short sentences more typical of English, the names of the characters, names of places left in English, the setting in Chicago, and the use of the term 2BR02B.

As we have observed in previous experiments, the main give away that MT was a translation relies in the quality of the language, while in the other conditions, the main give away is the setting of the story.

*Final Comments*

In this section, the comments were varied, and they did not differ noticeably by reading condition. The readers commented that they either liked or not liked the story, some found it "fascinating and intriguing" while others found it "unpleasant", "terribly dark text". They also mentioned whether they like or dislike Science Fiction or dystopian literature. Many were intrigued by the experiment and wanted to know the results. One NL reader in the MT version commented, "I happen to know that this is a story by Kurt Vonnegut. But I can hardly imagine that someone has translated his story in the way presented before. This translation really does not do justice to his authorship" which happens to touch upon an issue often discussed among translators and scholars about the role of MT in promoting literature in unusual language combinations. Can MT really do justice to authorship?

*Conclusions*

In the second part of the CREAMT project, the data obtained supports partially the conclusions obtained in a pilot experiment (Guerberof-Arenas and Toral 2020). CA users reading a text processed with MT have different reading experiences, and they are more engaged when they read HT. In this experiment, the HT version also scores significantly higher than PE in many of the scales. This is not surprising, as the HT version had scored higher in creative shifts and lower in errors when comparing it with the PE and MT versions during the review (Guerberof-Arenas and Toral 2022). Moreover, the CA reviewers praised many of the solutions found in the HT version (even if the PE version was post-edited by the same two translators).

However, when analysing the NL group, the results differ. They show higher values when reading the original ST[30] or the PE version than when reading MT or HT. For many categories, the reading condition does not show any significant difference even if PE tends to score higher. These results seem odd since all NL texts were rated by professional reviewers and they found that the PE version, when compared to HT, was too close to the English and seemed to be translated by amateurs or novel translators and, in addition, it contained far more errors than the HT version. We hypothesize that the cause for these results could be the habituation of NL readers to read native English-speaking authors, and consequently they would favour a translation closer to the ST, in this case PE, than a more creative alternative in Dutch. Of course, this hypothesis would need to be tested with a different ST and source language. Overall, NL readers have a fuller reading experience when reading in the original language.

It must be said that, according to the reviewers, the CA translators performed better than the NL counteparts and this could also be the reason why the NL readers were not as impressed by the HT translation as the CA readers. The only category where NL readers rated HT significantly higher was in the Payment category, but only after being debriefed as to the nature of the translation. This means that readers do place a money value on the work done by professional translators, even if this was not reflected on the other scales.

Further, the participants found the quality of the CA MT system significantly higher than the NL system. It could be the case that the quality was indeed different even if automatic metrics did not reflect this, the NL translators also score the quality lower than the CA translators, but this could also be related to the higher reading patterns of the NL readers. The NL participants rated all the categories lower than the CA participants, so it could be that they were more demanding readers.

---

[30] This condition was not present in CA.

Regardless of the target language, we have also confirmed that when looking into narrative engagement of fictional pieces, the category *Narrative understanding*, and partially *Attentional focus*, are the ones most affected when using MT. While others, such as *Narrative presence*, *Emotional engagement* and *Visual imagery* are not significantly different in any condition. This is also replicated when looking at the comprehension questions devised for this story. However, readers commented on several critical issues related to language when reading MT, that might negatively impact, if used without editing, comprehension and the opinion of the author.

*Funding information*

This project has received funding from the European Union's Horizon 2020 research and innovation programme under the Marie Skłodowska-Curie grant agreement No. 890697.

*Acknowledgements*

We would like to thank the translators Carlota Gurt, Yannick Garcia Porres, Núria Molines Galarza, Josep Marco Borillo, Scheherezade Surià, Theo Schoemaker, Roos van de Wardt, Linda Broer and Leen van de Broucke, and the annotators Tia Nutters and Gerrit Bayer-Hohenwarter for their crucial contribution to this study. We would also like to thank all readers that contributed to this project section who on many occasions waved the offered payment. Thank you for making this possible!

Translations in Multilingual Online Reading Communities: Deriving Cognitive-Evaluative Templates from Big Data'. *Translation, Cognition & Behavior* 4.2 (December): 147–86. https://doi.org/10.1075/tcb.00060.kot.

Kruger, Haidee. 2013. 'Kruger: Child and Adult Readers' Processing of Foreign Elements in Translated South African Picturebooks'. *Target. International Journal of Translation Studies* 25 (2): 180–227. https://doi.org/10.1075/target.25.2.03kru.

Kruger, Jan-Louis. 2018. 'Eye Tracking in Audiovisual Translation Research'. In *The Routledge Handbook of Audiovisual Translation*, edited by Luis Pérez-González, 1st ed., 350–66. London, New York: Routledge. https://doi.org/10.4324/9781315717166-22.

Kuijpers, Moniek M., Frank Hakemulder, Ed S. Tan, and Miruna M. Doicaru. 2014. 'Exploring Absorbing Reading Experiences: Developing and Validating a Self-Report Scale to Measure Story World Absorption'. *Scientific Study of Literature* 4 (1): 89–122. https://doi.org/10.1075/ssol.4.1.05kui.

Mangen, Anne, and Don Kuiken. 2014. 'Lost in an IPad: Narrative Engagement on Paper and Tablet'. *Scientific Study of Literature* 4 (2): 150–77. https://doi.org/10.1075/ssol.4.2.02man.

Nuland, Sherwin B. 1995. *How We Die: Reflections of Life's Final Chapter, New Edition*. 1st edition. New York: Vintage.

Orrego-Carmona, David. 2018. 'Audiovisual Translation and Audience Reception'. In *The Routledge Handbook of Audiovisual Translation*, edited by Luis Pérez-González, 1st ed., 367–82. London, New York: Routledge. https://doi.org/10.4324/9781315717166-23.

Ortiz Boix, Carla. 2016. 'Implementing Machine Translation and Post-Editing to the Translation of Wildlife Documentaries through Voice-over and off-Screen Dubbing'. *TDX (Tesis Doctorals En Xarxa)*. Ph.D. Thesis, Barcelona: Universitat Autònoma de Barcelona. http://www.tdx.cat/handle/10803/400020.

Stasimioti, Maria, and Dr Vilelmini Sosoni. 2022. 'Creative Texts Translation vs Post-Editing: A Qualitative Study of the Product Quality, the Translators' Perception and Audience's Reception'. In *Workshop on Creativity and Technology*.37

*Authors' Address*

**Ana Guerberof-Arenas**

https://orcid.org/0000-0001-9820-7074

Computational Linguistics Group

Center for Language and Cognition

Faculty of Arts

University of Groningen

The Netherlands

a.guerberof.arenas@rug.nl

*Antonio Toral*

https://orcid.org/0000-0003-2357-2960

Computational Linguistics Group

Center for Language and Cognition

Faculty of Arts

University of Groningen

The Netherlands

a.toral.ruiz@rug.nl